\documentclass[a4paper,fleqn]{cas-dc}

\usepackage[sort&compress]{natbib}
\usepackage{caption}
\usepackage{array}
\usepackage{booktabs}
\usepackage{multicol}
\usepackage{multirow}
\usepackage{fontawesome}
\usepackage{xcolor}
\usepackage{algorithm}
\usepackage{algpseudocode}
\usepackage{amsmath}
\usepackage{amssymb}

\def\tsc#1{\csdef{#1}{\textsc{\lowercase{#1}}\xspace}}
\tsc{WGM}
\tsc{QE}

\begin{document}
\let\WriteBookmarks\relax
\def\floatpagepagefraction{1}
\def\textpagefraction{.001}

\title[mode = title]{Hierarchical Prototype-based Domain Priors for Multiple Instance Learning in Multimodal Histopathology Analysis}

\author[1]{Xuemei Qiu}
\fnmark[1]
\ead{xuemei.qiu@fafu.edu.cn}

\affiliation[1]{organization={College of Computer and Information Science, Fujian Agriculture and Forestry University},
            city={Fuzhou},
            postcode={350002}, 
            country={China}}

\affiliation[2]{organization={College of Future Technology, Fujian Agriculture and Forestry University}, 
            city={Fuzhou},
            postcode={350002}, 
            country={China}}
            
\affiliation[3]{organization={Digital Fujian Institute of Agricultural Big Data, Fujian Agriculture and Forestry University},
            city={Fuzhou},
            postcode={350002}, 
            country={China}}

\affiliation[4]{organization={Department of Pathology, Clinical Oncology School of Fujian Medical University, and Fujian Cancer Hospital},
            city={Fuzhou},
            postcode={350014}, 
            country={China}}

\author[1]{Dawei Fan}
\ead{dawei.fan@fafu.edu.cn}
\fnmark[1]

\author[1]{Yebin Huang}
\ead{12411093021@fafu.edu.cn}

\author[4]{Yanping Chen}
\ead{chenyanping@fjzlhospital.com}

\author[1,2,3]{Lifang Wei}
\cormark[1]
\ead{weilifang0028@fafu.edu.cn}

\fntext[1]{Equal contribution.}
\cortext[1]{Corresponding author at: College of Computer and Information Science, Fujian Agriculture and Forestry University, Fuzhou, 350002, China.}


\begin{abstract}
Digital pathology has fundamentally altered diagnostic workflows by enabling the computational analysis of gigapixel Whole Slide Images (WSIs), yet effectively deciphering their complex tumor microenvironments remains a formidable challenge. Existing Multiple Instance Learning (MIL) frameworks typically treat Whole Slide Images as unstructured bags of patches, discarding critical morphological semantics and spatial geometry. This lack of inductive bias often leads to overfitting on background noise and fails to align visual features with high-level diagnostic knowledge. To overcome these limitations, we propose the Hierarchical Prototype-based Domain Priors (HPDP) framework, a unified multimodal approach for joint histopathology diagnosis and prognosis. HPDP mitigates the data-driven "black box" issue by introducing a Morphologically Anchored Prototype System (MAPS), which anchors learning to interpretable morphological clusters, and a Sinusoidal Positional Encoder (SPE) to explicitly model tissue architecture. Furthermore, we bridge the semantic gap via a Hierarchical Cross-Modal Alignment (HCMA) module, using Large Language Model (LLM)-generated descriptions to contextually refine visual representations. Extensive experiments across seven cancer cohorts demonstrate that HPDP consistently achieves state-of-the-art performance with superior robustness and interpretability.
\end{abstract}


\maketitle

\section{Introduction}
Histopathological examination serves as the clinical gold standard for cancer diagnosis, tumor subtyping, and prognostic assessment \citep{chen2024towards,vorontsov2024foundation}. With the digitization of histology into Whole Slide Images (WSIs), the field has witnessed a paradigm shift toward computational pathology where deep learning algorithms promise to alleviate the workload of pathologists while enhancing diagnostic objectivity \citep{echle2021deep,madabhushi2016image,campanella2019clinical}. Given the gigapixel resolution of WSIs and the prohibitive cost of pixel-level annotation, Multiple Instance Learning (MIL) has established itself as the dominant framework by conceptualizing a slide as a bag of patch-level instances to learn robust slide-level representations from weak supervision \citep{xu2024whole,campanella2019clinical,ilse2018attention,li2021dual,lu2021data,shao2021transmil,zhang2022dtfd}. This approach has fueled significant advances in automated tumor detection and survival risk stratification without requiring dense manual segmentation \citep{vorontsov2024foundation}.

\begin{figure*}
    \centering
    \includegraphics[width=1\linewidth]{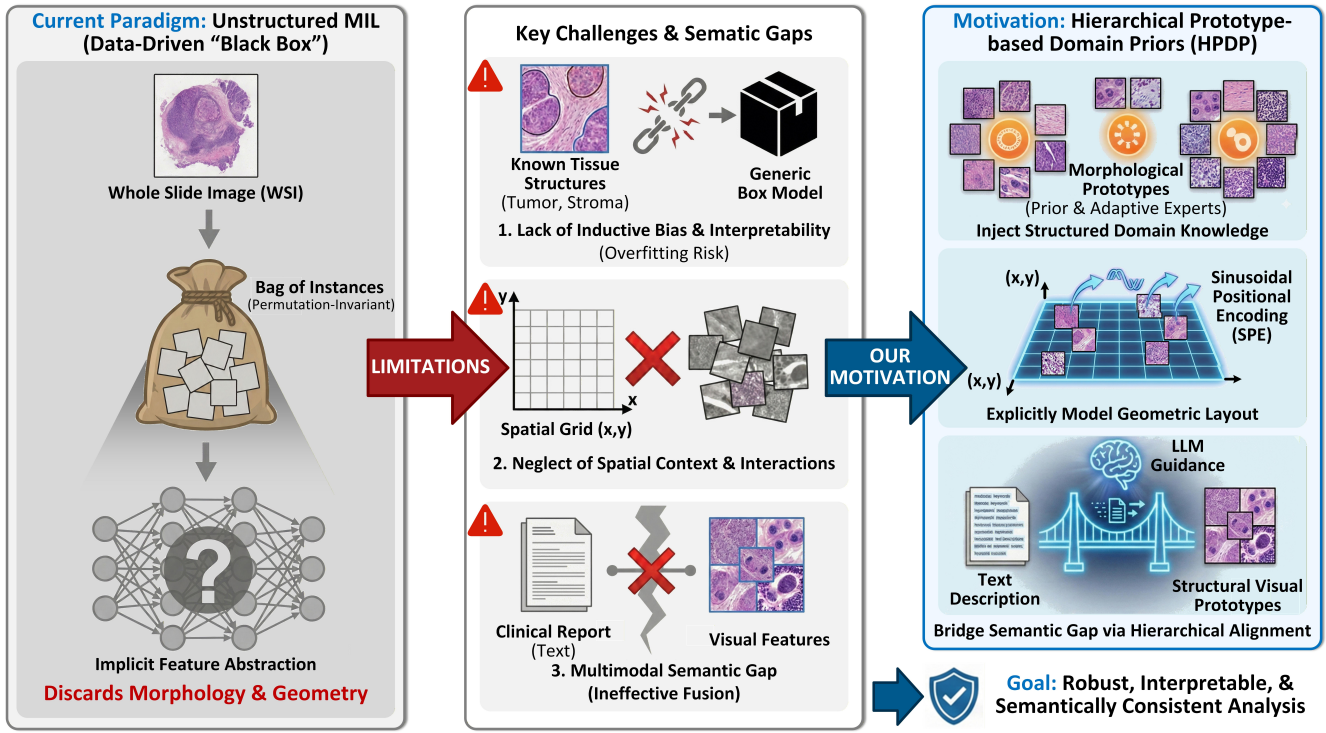}
    \caption{Illustration of the research motivation and the proposed Hierarchical Prototype-based Domain Priors framework. The diagram depicts the limitations of the conventional unstructured MIL paradigm (Left), identifies the key challenges regarding inductive bias, spatial context, and multimodal gaps (Middle), and presents the proposed HPDP approach (Right) which integrates morphological prototypes, geometric encoding, and LLM guidance.}
    \label{fig1}
\end{figure*}

Despite these strides, current MIL paradigms face intrinsic limitations rooted in their reliance on purely data-driven feature abstraction \citep{shi2023structure}. Most existing approaches, as shown in Fig. \ref{fig1}, treat the WSI as an unstructured collection of patches and effectively discard the rich domain knowledge inherent in tissue morphology where pathologists rely on distinct entities like tumor nests, stroma, and infiltrating lymphocytes to form a diagnosis \citep{campanella2019clinical,ilse2018attention,shao2021transmil}. Conventional attention-based or Transformer-based MIL models are often tasked with rediscovering these well-established patterns from scratch without explicit guidance, a lack of inductive bias that increases the risk of overfitting to background noise particularly in limited-data regimes \citep{lu2024visual,zhang2024attention}. This issue is further compounded by the neglect of the tissue microenvironment's geometric layout as treating slides as permutation-invariant bags fails to capture the spatial interactions and relative positioning of cells that are critical for accurate prognosis \citep{hou2022h}.

Beyond the visual domain, the integration of multimodal context represents a largely untapped frontier in histopathology \citep{huang2023visual,lu2024visual}. While clinicians synthesize visual cues with diagnostic background knowledge to assess a patient, computational models rarely benefit from such high-level semantic guidance. The emergence of Large Language Models (LLMs) offers an unprecedented opportunity to bridge this gap by generating coherent clinical descriptions from histological data, yet effectively fusing these linguistic insights with visual features remains non-trivial \citep{lu2024multimodal,10.5555/3666122.3667776}. Naive concatenation of modalities often fails to bridge the semantic gap and leaves the visual representation unmodulated by the specific clinical context \citep{zhang2023text,watawana2024hierarchical}.

To navigate these challenges, we propose the Hierarchical Prototype-based Domain Priors (HPDP) framework, a novel multimodal approach designed to inject structured morphological and semantic knowledge into the MIL pipeline. Departing from the black-box nature of traditional aggregators, we introduce a Morphologically Anchored Prototype System (MAPS) that explicitly anchors the learning process using "Prior Experts" initialized via unsupervised clustering to represent canonical morphological categories, while simultaneously employing "Adaptive Experts" to mine subtle and task-specific discriminative patterns. Complementing this morphological foundation, we incorporate a Sinusoidal Positional Encoder (SPE) that maps the spatial coordinates of each patch into the feature space to enable the perception of relative distances within the tissue architecture. To further bridge the visual-linguistic divide, our framework integrates a Hierarchical Cross-Modal Alignment (HCMA) module where LLM-generated clinical text functions as a semantic controller through a two-stage process of global semantic modulation and structural context propagation. This comprehensive design ensures that the resulting slide representation is both visually discriminative and semantically consistent, a capability demonstrated by our extensive experiments across multiple cancer cohorts where HPDP consistently achieves state-of-the-art performance in both diagnosis and prognosis tasks.

In summary, this work makes the following contributions:

\begin{itemize}
\item We propose the Hierarchical Prototype-based Domain Priors framework for joint histopathology diagnosis and prognosis. This approach uniquely synergizes K-means-derived morphological priors with LLM-generated semantic guidance, effectively bridging the gap between data-driven feature learning and expert-defined domain knowledge.
\item We design a novel Morphologically Anchored Prototype System that disentangles slide-level representation learning. By synergizing fixed "Prior Experts" with learnable "Adaptive Experts", coupled with a Sinusoidal Positional Encoder, the model effectively captures both morphological semantics and geometric context.
\item We introduce the Hierarchical Cross-Modal Alignment module to bridge the visual-linguistic gap. By employing global semantic modulation and structural context propagation, it leverages LLM-generated descriptions as anchors to contextually refine visual features for diagnostically relevant attention.
\item We conduct extensive experiments on seven diverse datasets encompassing multiple cancer types. The results demonstrate that HPDP consistently achieves state-of-the-art performance in both classification and survival tasks, exhibiting superior robustness and interpretability compared to existing single-modal and multimodal baselines.
\end{itemize}

\section{Related work}
In this section, we review the literature relevant to our framework from three perspectives: Multiple Instance Learning strategies in WSI analysis , the integration of prototype learning with domain priors , and recent advancements in multimodal vision-language learning for pathology.
\subsection{Multiple Instance Learning in WSI Analysis}
Multiple Instance Learning (MIL) is a widely recognized approach to the classification of Whole Slide Images (WSIs), especially in scenarios with weak supervision \citep{song2023artificial}. In this paradigm, each WSI is represented as a "bag" $\mathcal{B} = \{ x_1, x_2, \dots, x_N\}$ consisting of $N$ image patches at the instance level, while only a single slide-level label $\mathcal{Y} \in \{0, \dots, C-1\}$ is provided. Existing MIL approaches can generally be divided into two broad categories: (1) Explicit modeling, where the slide label is predicted by directly aggregating instance-level predictions, i.e., $\mathcal{\hat{Y}} = \text{pool}\{\hat{y}_1, \hat{y}_2, \dots, \hat{y}_N\}$, using techniques such as mean or max pooling \citep{campanella2019clinical, zhang2022dtfd}. (2) Implicit representation learning, where a comprehensive slide-level representation $z = f(\mathcal{B})$ is learned by aggregating instance embeddings, followed by classification using recurrent neural networks (RNNs) \citep{campanella2019clinical} or attention mechanisms \citep{ilse2018attention}.

In the latter category, attention-based MIL methods have proven highly effective by learning adaptive instance weights to mitigate the inductive bias in complex histopathological patterns. Architectural innovations have further enhanced this capability; for instance, \cite{lu2021data} introduced an instance-level clustering branch to separate distinctive features, while TransMIL \citep{shao2021transmil} utilized self-attention mechanisms to capture long-range inter-instance dependencies. More recent strategies have focused on optimizing the learning trajectory, such as improving instance selection stability \citep{yu2023bayesian,li2023task}, enhancing feature representation via contrastive sampling \citep{wang2022scl}, or ensuring robustness against dominant background patches \citep{zhang2024attention}.

\subsection{Prototype Learning and Domain Priors}
Integrating prototype learning into histopathology has recently gained traction as a means to enhance model interpretability and handle intra-class heterogeneity \citep{liu2024pamil,NEURIPS2024_760341ad,liu2025prototype}. Emerging works from the past two years have pushed this frontier significantly. For instance, \cite{NEURIPS2024_760341ad} proposes a heterogeneous graph framework that explicitly differentiates tissue types  within the tumor microenvironment. By decoupling representation learning into a structure view for spatial context and a histology view for prototype-based abstraction, it effectively integrates pathological domain knowledge. Similarly, \cite{liu2025prototype} introduces a prototype-guided cross-modal mechanism designed to enhance survival prediction. It aligns query features with learnable prognostic prototypes to distill risk-related semantics from established patterns. Another notable approach, \cite{WANG2024103252} employs a dual-stream multi-dependency graph architecture. By modeling WSIs via distinct graphs based on spatial proximity and morphological affinity, it captures both local geometric interactions and global semantic consistencies to boost performance.

However, despite these advancements, a key limitation persists: most existing prototype-based methods learn prototypes in a purely latent space or optimize them solely through end-to-end task supervision \citep{rymarczyk2021kernel}. They often lack explicit constraints anchored in biological reality, meaning the learned "prototypes" may drift towards background noise or fail to represent canonical morphological entities \citep{fan2025ipmmg}. To address this limitation, we propose the MAPS. Unlike previous latent approaches, MAPS explicitly initializes "Prior Experts" using K-means clustering of tissue phenotypes, enforcing a structured domain prior. This design ensures that the model respects established morphological categories while simultaneously employing "Adaptive Experts" to mine subtle, task-specific patterns that predefined clusters might miss.

\subsection{Multimodal Vision-Language Learning in Pathology}
The intersection of computational pathology and LLMs represents the cutting edge of current research, moving beyond purely visual analysis to incorporate clinical semantic guidance \citep{shi2024vila}. Recent frameworks have introduced sophisticated mechanisms to align these modalities. Notably, \cite{shi2024vila} proposes a dual-scale vision-language learning framework. It introduces a bag-level alignment to capture global diagnostic concepts and an instance-level alignment to ground fine-grained morphological details, effectively enhancing feature discriminability. Similarly, \cite{Lu_2023_CVPR} explores the potential of transferring pre-trained vision-language models to histopathology for zero-shot classification. By leveraging carefully engineered text prompts to represent pathological categories, it demonstrates the efficacy of linguistic supervision without the need for task-specific fine-tuning. Furthermore, the release of large-scale curated datasets like Quilt-1M \citep{10.5555/3666122.3667776} has facilitated the training of foundation models such as CONCH and PLIP. These models utilize contrastive learning objectives on massive image-text pairs to learn robust, semantically aligned representations that generalize well across diverse histopathological tasks.

Nevertheless, a significant "semantic gap" remains in how these modalities interact in specific downstream tasks \citep{chen2021multimodal}. Most current state-of-the-art methods rely on global contrastive losses or simple cross-attention mechanisms, which align features broadly but fail to use the patient-specific clinical description as a precise "controller" for visual attention \citep{shi2024vila,zhang2023text}. They often treat text as a passive auxiliary signal rather than an active modulator of the visual latent space. Differently, our HCMA module bridges this gap by employing a two-stage modulation process. We utilize LLM-generated text to dynamically recalibrate visual prototypes via feature-wise modulation and structural context propagation, ensuring that the model's focus is explicitly guided by high-level diagnostic reasoning rather than low-level visual statistics alone.

\section{Methodology}
In this section, we introduce the HPDP framework, which is a new multimodal Multiple Instance Learning approach for joint histopathology diagnosis and prognosis. An overview of the HPDP framework and a detailed description of its key pipelines and modules, including the Information Processing Pipeline, Feature Feed-forward, and Cross-Modal Representation Learning, are provided below.

\begin{figure*}
    \centering
    \includegraphics[width=1\linewidth]{fig2.jpg}
    \caption{The architecture of the proposed HPDP framework. (a) The pipeline processes WSI patches to generate K-means-based domain priors and LLM-based clinical text. (b) Visual features are fused with SPE for spatial awareness, while text features are extracted via a frozen encoder. (c) The MAPS aggregates features using both Prior and Adaptive Experts. The resulting prototypes are aligned with text embeddings via the HCMA module to predict diagnostic and prognostic outcomes.}
    \label{fig2}
\end{figure*}
\subsection{Overview of the framework}
As shown in Fig. \ref{fig2}, the framework of the proposed HPDP comprises of the information processing pipeline, feature feed-forward, and cross-modal representation learning. Following these steps, the unstructured histological images, spatial coordinates, and clinical text description can be transformed into a unified high-dimensional feature space, enabling the model to perform joint diagnosis and prognosis efficiently.

\textit{Information Processing Pipeline:} As illustrated in Fig. \ref{fig2}(a), the information processing pipeline constitutes the foundation of our framework, comprising three parallel procedures: WSI preprocessing, domain prior construction, and clinical text generation. 
Initially, we employ an automated tissue detection algorithm to differentiate the foreground from the background of each WSI \citep{otsu1975threshold}. The identified tissue regions are then tessellated into numerous non-overlapping fixed-size patches. Subsequently, these patches are encoded by a pretrained ResNet-50 encoder \citep{He2015} into a sequence of patch-level features, denoted as a bag of instances $F = \{f_1, f_2, \dots, f_N\} \in \mathbb{R}^{N \times 1024}$. Crucially, during the cropping process, the spatial coordinates $(x, y)$ of each patch are recorded to preserve the geometric layout of the tissue microenvironment for subsequent positional encoding. A distinctive aspect of our pipeline is the generation of domain priors. We apply K-means clustering on the patch population to identify representative morphological centroids (e.g., distinguishing categories such as vessels, cells, and stroma). These centroids function as supervised prototypes $\mathcal{P}_{prior} \in \mathbb{R}^{\mathcal{K} \times D}$ (where $\mathcal{K}$ denotes the number of morphological clusters) for the subsequent expert prior loss, thereby effectively injecting structured domain knowledge into the model. 
Simultaneously, to efficiently utilize semantic context, we employ a LLM \citep{OpenAI2023GPT4}. By processing the downsampled WSIs, the LLM generates coherent text descriptions, providing high-level semantic anchors for the visual features.

\textit{Feature Feed-forward:} Subsequent to the initial processing, as illustrated in Fig. \ref{fig2}(b), we conduct feature feed-forward operations to embed the raw data into high-dimensional feature vectors. This stage processes the visual, spatial, and textual modalities respectively to generate task-agnostic representations. Regarding spatial representation, to explicitly model the geometric layout of the tissue microenvironment, we map the normalized coordinates $(x_i, y_i)$ of each patch into high-dimensional embeddings via SPE. The SPE applies sine and cosine functions of varying frequencies, enabling the model to attend to the relative distances between patches. Subsequently, regarding the visual features, we employ a Multi-Layer Perceptron (MLP) to project them into a target dimension $D$, yielding the projected features $\mathcal{F}$. Consequently, $\mathcal{F}$ is fused with the spatial embeddings via element-wise addition to form the location-aware slide features, denoted as $\mathcal{H} = \{h_1, h_2, \dots, h_N\} \in \mathbb{R}^{N \times D}$. This fusion ensures that each instance concurrently carries both morphological and geometric information. Simultaneously, the clinical description generated by the LLM is tokenized and passed through a frozen BioBERT text encoder \citep{lee2020biobert}, yielding a global semantic embedding $T \in \mathbb{R}^{1 \times D}$ that encapsulates the patient's diagnostic background information.

\textit{Cross-Modal Representation Learning:} In this phase, the extracted slide features $F$ and text embeddings $T$ are fed into the core network for representation learning. As illustrated in Fig. \ref{fig2}(c), the system comprises the MAPS and the HCMA modules. The MAPS functions as the central aggregation engine, transforming the slide features into compact bag-level representations via a dual-path mechanism. It utilizes Prior Experts to capture established domain-invariant patterns, while simultaneously employing Adaptive Experts to discover task-specific nuances through end-to-end learning. 
Subsequently, the HCMA module dynamically modulates these aggregated prototypes using the global text feature $T$. By hierarchically aligning visual representations with clinical semantic guidance, the model ensures that the histopathological features are contextually refined. Eventually, the optimized features are pooled and forwarded to specific task heads to predict the diagnostic class probabilities for tumor subtyping or the hazard rates for survival analysis, thereby realizing a comprehensive assessment of the patient's condition.

\subsection{Sinusoidal Positional Encoder (SPE)}

In histopathology, the spatial arrangement of tissue patches contains critical diagnostic information. Standard MIL approaches often treat the slide as an unordered bag of instances, thereby discarding this structural context \citep{chen2021whole}. To address this, we introduce a fixed SPE to explicitly model the geometric layout of the WSI.

Given the normalized spatial coordinates $p_i = (x_i, y_i)$ for the $i$-th patch and a target embedding dimension $D$, we independently encode the horizontal ($x$) and vertical ($y$) positions. The dimension $D$ is split equally, allocating $D/2$ dimensions to each coordinate axis. 

Inspired by the positional encoding in Transformers, we utilize sinusoidal functions of varying frequencies to map the scalar coordinates into high-dimensional vectors. For a given coordinate $c \in \{x_i, y_i\}$, the encoding is defined as:
\begin{equation}
    \begin{aligned}
        PE(c, 2j) &= \sin\left(c \cdot 10000^{-4j/D}\right) \\
        PE(c, 2j+1) &= \cos\left(c \cdot 10000^{-4j/D}\right)
    \end{aligned}
\end{equation}
where $j \in [0, D/4 - 1]$ denotes the frequency index. The term $10000^{4j/D}$ corresponds to the wavelength, forming a geometric progression from $2\pi$ to $10000 \cdot 2\pi$. This formulation allows the model to attend to relative positions by learning to recognize linear functions of the input coordinates.

Finally, the embeddings for the $x$ and $y$ axes are concatenated to form the final spatial feature vector $S_i \in \mathbb{R}^D$:
\begin{equation}
    \begin{split}
        S_i &= \text{Concat}\big[PE(x_i), PE(y_i)\big] \\
            &= \big[\sin(x_i), \cos(x_i), \sin(y_i), \cos(y_i)\big]
    \end{split}
\end{equation}
To preserve the spatial context within the histological representation, these spatial embeddings are fused with the projected visual features $\mathcal{F}_i$ via element-wise addition. This yields the final location-aware slide features, denoted as $\mathcal{H} = \{h_1, \dots, h_N\}$, where $h_i = \mathcal{F}_i + S_i$. This set $\mathcal{H}$ serves as the input for the subsequent MAPS aggregation.

\subsection{Morphologically Anchored Prototype System (MAPS)}
To ameliorate the challenge of discerning salient instances from vast gigapixel WSIs, we engineer the MAPS. This design is motivated by the need to reconcile the trade-off between structural stability (often provided by unsupervised clustering) and task-specific plasticity (driven by supervised attention). Consequently, MAPS disentangles slide-level representation learning into two complementary subspaces. Crucially, we tailor the interaction mechanisms to their geometric roles: employing magnitude-invariant Cosine Similarity to capture semantic directionality, and magnitude-sensitive Dot Product to model feature intensity. 

\textbf{Semantic Routing via Manifold Alignment.}
The first pathway injects structural inductive biases into the network by leveraging $\mathcal{K}_{sup}$ Prior Experts. These experts are instantiated as learnable parameters, denoted as $\mathcal{P}_{prior} \in \mathbb{R}^{\mathcal{K}_{sup} \times D}$. To ensure they capture established morphological patterns (e.g., tumor, stroma), their optimization trajectory is guided by the K-means derived centroids (Export Prototypes) via an auxiliary supervision loss. This mechanism anchors the experts to meaningful domain semantics while retaining the flexibility of gradient-based tuning. To focus strictly on semantic directionality, we project both inputs and experts onto a unit hypersphere, deriving routing attention via cosine similarity:
\begin{equation}
    \mathcal{A}_{prior} = \text{Softmax}\left(\frac{1}{\tau} \cdot \frac{\mathcal{H} \mathcal{P}_{prior}^\top}{\|\mathcal{H}\|_2 \|\mathcal{P}_{prior}\|_2}\right) \in \mathbb{R}^{N \times \mathcal{K}_{sup}}
\end{equation}
where $\|\cdot\|_2$ denotes the $L_2$ norm, and $\tau$ is the temperature coefficient. This process effectively ``routes'' patches to their semantically nearest expert prototypes.

\textbf{Latent Feature Mining via Adaptive Interaction.}
Complementary to the supervised priors, the second pathway introduces $\mathcal{K}_{free}$ Adaptive Experts, denoted as $\mathcal{P}_{adapt} \in \mathbb{R}^{\mathcal{K}_{free} \times D}$, also initialized as free learnable parameters but optimized without external morphological constraints. These experts are tasked with mining subtle, task-specific discriminative patterns that may lie outside the pre-defined clusters. Here, we employ a raw dot-product interaction to preserve the magnitude information, which reflects the activation intensity of the features:
\begin{equation}
    \mathcal{A}_{adapt} = \text{Softmax}\left(\frac{\mathcal{H} \mathcal{P}_{adapt}^\top}{\tau}\right) \in \mathbb{R}^{N \times \mathcal{K}_{free}}
\end{equation}
This unconstrained interaction allows the model to dynamically construct new prototypes during end-to-end training.

\textbf{Unified Attentive Aggregation.}
To synthesize a comprehensive slide representation, we concatenate the attention landscapes from both pathways, yielding a unified weight matrix $\mathcal{A}_{total} = [\mathcal{A}_{prior}, \mathcal{A}_{adapt}] \in \mathbb{R}^{N \times (\mathcal{K}_{sup} + \mathcal{K}_{free})}$. The final aggregated prototypes $\mathcal{M} \in \mathbb{R}^{\mathcal{K} \times D}$ are computed via a linear combination of the input instances weighted by $\mathcal{A}_{total}$:
\begin{equation}
    \mathcal{M} = \mathcal{A}_{total}^\top \mathcal{H}
\end{equation}
By explicitly coupling domain knowledge with adaptive learning, MAPS generates a representation that is both semantically interpretable and highly discriminative for downstream tasks.

\subsection{Hierarchical Cross-Modal Alignment (HCMA)}

While the MAPS effectively aggregates morphological patterns into a compact prototype set $\mathcal{M} \in \mathbb{R}^{\mathcal{K} \times D}$, these representations remain strictly in the visual domain. To bridge the semantic gap and explicitly incorporate clinical guidance, we introduce the HCMA module. This module aligns visual and textual modalities through a cohesive two-stage process: Global Semantic Modulation and Structural Context Propagation.

\textbf{Stage 1: Global Semantic Modulation.}
First, to inject the global linguistic context into the visual latent space, we employ a Feature-wise Linear Modulation mechanism. This performs a channel-wise recalibration of the visual prototypes conditioned on the patient-specific text embedding $T$. Specifically, a parameter generation network maps the text $T$ to a set of affine transformation parameters: a scaling factor $\boldsymbol{\gamma} \in \mathbb{R}^{1 \times D}$ and a shifting factor $\boldsymbol{\beta} \in \mathbb{R}^{1 \times D}$. The modulated features $\mathcal{H}_{mod}$ are computed via a residual structure:
\begin{equation}
    \mathcal{H}_{mod} = \text{LayerNorm}\big(\mathcal{M} + (\boldsymbol{\gamma}(T) \odot \mathcal{M} + \boldsymbol{\beta}(T))\big)
\end{equation}
where $\odot$ denotes the element-wise Hadamard product. This operation allows the clinical description to selectively suppress irrelevant visual channels while amplifying those pertinent to the diagnosis.

\textbf{Stage 2: Structural Context Propagation.}
While the modulation injects semantics globally, it operates point-wise. To stabilize the representation and preserve the structural integrity of the original visual prototypes, we employ a Multi-Head Attention (MHA) mechanism for feature propagation. Crucially, the original prototypes $\mathcal{M}$ serve as the Query, while the semantically modulated features $\mathcal{H}_{mod}$ act as both Key and Value. This design queries the text-enhanced features using the original visual structure as a guide:
\begin{equation}
    \mathcal{M}_{final} = \text{LayerNorm}\left(\mathcal{M} + \text{MHA}(\mathcal{M}, \mathcal{H}_{mod}, \mathcal{H}_{mod})\right)
\end{equation}
By adding the propagated information back to the original $\mathcal{M}$ via a residual connection, the final representations $\mathcal{M}_{final}$ retain robust visual characteristics while being deeply aligned with high-level clinical semantic guidance.

\subsection{Objective Function}

The proposed HPDP framework is optimized in an end-to-end manner. The total objective function $\mathcal{L}_{total}$ is composed of a task-specific loss $\mathcal{L}_{task}$ (either for classification or survival analysis) and an auxiliary prototype supervision loss $\mathcal{L}_{proto}$ to enforce the structural domain priors.

\textbf{Task-Specific Loss.}
For classification tasks, we employ the standard Cross-Entropy loss to minimize the divergence between the predicted probability distribution and the ground truth labels:
\begin{equation}
    \mathcal{L}_{cls} = -\frac{1}{B} \sum_{i=1}^{B} \sum_{c=1}^{C} y_{i,c} \log(\hat{y}_{i,c})
\end{equation}
where $B$ is the batch size, $C$ is the number of classes, and $\hat{y}$ denotes the predicted probability.

For survival analysis, we adopt the Cox Proportional Hazards Loss (Negative Log Partial Likelihood). Given a batch of patients sorted by survival time in descending order, for the $i$-th patient with risk score $\hat{h}_i$ and event status $\delta_i$ ($\delta_i=1$ indicates an observed event, $\delta_i=0$ indicates censoring), the loss is formulated as:
\begin{equation}
    \mathcal{L}_{surv} = -\frac{1}{B} \sum_{i: \delta_i = 1} \left( \hat{h}_i - \log \left( \sum_{j \in \mathcal{R}(t_i)} \exp(\hat{h}_j) + \epsilon \right) \right)
\end{equation}
where $\mathcal{R}(t_i) = \{j \mid t_j \ge t_i\}$ represents the set of patients at risk at time $t_i$, and $\epsilon$ is a small constant for numerical stability.

\textbf{Prototype Supervision Loss.}
To explicitly guide the Prior Experts $\mathcal{P}_{prior}$ in the MAPS to capture meaningful morphological semantics, we introduce an auxiliary supervision loss. We align the learnable prior experts with the fixed Export Prototypes $\mathcal{P}_{teacher}$ using a contrastive alignment mechanism. Specifically, we maximize the cosine similarity between the $k$-th learnable expert and its corresponding teacher prototype while minimizing its similarity to other prototypes. This is implemented via a Cross-Entropy loss over the similarity matrix:
\begin{equation}
    \mathcal{L}_{proto} = -\frac{1}{\mathcal{K}_{sup}} \sum_{k=1}^{\mathcal{K}_{sup}} \log \frac{\exp(\text{sim}(\mathbf{p}_k, \mathbf{t}_k) / \tau)}{\sum_{j=1}^{\mathcal{K}_{sup}} \exp(\text{sim}(\mathbf{p}_k, \mathbf{t}_j) / \tau)}
\end{equation}
where $\mathbf{p}_k \in \mathcal{P}_{prior}$ and $\mathbf{t}_k \in \mathcal{P}_{teacher}$ are the $k$-th learnable and teacher prototypes respectively, $\text{sim}(\cdot, \cdot)$ denotes cosine similarity, and $\tau$ is the temperature.

\textbf{Total Objective.}
The final objective function is defined as the sum of the task loss and the prototype supervision loss:
\begin{equation}
    \mathcal{L}_{total} = \mathcal{L}_{task} + \lambda \mathcal{L}_{proto}
\end{equation}
where $\mathcal{L}_{task}$ switches between $\mathcal{L}_{cls}$ and $\mathcal{L}_{surv}$ depending on the downstream task, and $\lambda$ is a balancing hyperparameter.

\section{Experiment}

\subsection{Datasets}

To comprehensively evaluate the effectiveness and robustness of the HPDP framework, we conducted experiments on a diverse suite of histopathology datasets, comprising both public benchmarks and private in-house cohorts. These datasets cover a broad spectrum of tasks, ranging from tumor diagnosis and subtyping to survival risk prediction.

\textbf{Classification Datasets:}

    \textbf{Lung Cancer Cohort (LCEM).} We utilized an independent in-house dataset ($n=1291$) to perform molecular prediction of EGFR status in Lung Adenocarcinoma. The task is formulated as a binary classification problem to distinguish between Wild-type and Mutant EGFR genotypes.
    
    \textbf{TCGA-Lung.} Sourced from The Cancer Genome Atlas, this dataset comprises the two major non-small cell lung cancer subtypes ($n=1053$): Lung Adenocarcinoma (LUAD) and Lung Squamous Cell Carcinoma (LUSC). It serves as a standard benchmark for tumor subtyping classification, requiring the model to distinguish between these distinct histological patterns.
    
    \textbf{Camelyon16.} This public benchmark focuses on the detection of metastasis in sentinel lymph nodes of breast cancer, consisting of 399 WSIs (270 for training, 129 for testing). The task is designed as a binary classification problem to differentiate between normal slides and those containing metastases.
    
    \textbf{BRACS.} To evaluate performance in fine-grained scenarios, we employed the BReast CAncer Subtyping dataset ($n=547$). It targets a challenging multi-class classification task, aiming to distinguish between three core pathological categories: Benign, Atypical, and Malignant, providing an ideal platform for complex diagnosis.

\textbf{Survival Analysis Datasets:}

For prognostic assessment, we focused on predicting patient overall survival (OS) using three cohorts from TCGA and our large-scale in-house data.

\textbf{TCGA-LUAD.} This cohort consists of 528 samples from patients with Lung Adenocarcinoma. Beyond classification, it is utilized here to evaluate the model's capability in survival risk stratification for lung cancer.

\textbf{TCGA-COAD.} A cohort representing Colon Adenocarcinoma, consisting of 457 samples. It is employed to evaluate the model's ability to capture prognostic features from colorectal histology.

\textbf{TCGA-PAAD.} A challenging dataset for Pancreatic Adenocarcinoma with 209 samples. It is used to assess the model's performance on tumors characterized by complex stromal environments.

\textbf{LCEM.} Additionally, we utilized the private LCEM cohort ($n=1291$) for survival analysis. This real-world clinical dataset serves to validate the model's prognostic value in practical applications.

\subsection{Implementation Details}
\textbf{Data Preprocessing and Feature Engineering.} We utilized the Otsu thresholding algorithm to segment tissue regions from the background. From these regions, non-overlapping patches of size $256 \times 256$ pixels were extracted at $20\times$ magnification. Visual feature extraction was performed using a ResNet-50 backbone pre-trained on ImageNet \citep{He2015}. For the textual modality, clinical descriptions were synthesized using GPT-4.1 \citep{OpenAI2023GPT4}, and embeddings were extracted via a frozen BioBERT encoder \citep{lee2020biobert}.

\textbf{Experimental Settings.} All experiments were implemented using PyTorch on a single NVIDIA RTX 4090 GPU. We employed the Adam optimizer \citep{Kingma2015Adam} with a weight decay of $1\times 10^{-5}$. The learning rate was initialized at $3\times 10^{-4}$ and subsequently decayed to $1\times 10^{-4}$ using a scheduler. The model was trained for a maximum of 200 epochs, with an early stopping patience of 20 epochs. The balancing factor $\lambda$ in the total loss function was set to 0.1.

\textbf{Data Partitioning.} For the LCEM, TCGA-Lung, BRACS, TCGA-LUAD, TCGA-COAD, and TCGA-PAAD datasets, we adopted a random partitioning strategy (8:2 for train/test, with 20\% of training data for validation). For the Camelyon16 dataset, we strictly adhered to the official protocol.

\subsection{Evaluation Metrics}
To rigorously assess the performance of the proposed framework across different downstream tasks, we employ task-specific metrics.

\textbf{Classification Metrics.} For tumor diagnosis and subtyping tasks, we utilize three standard metrics: Accuracy (ACC), F1-Score, and the Area Under the Receiver Operating Characteristic Curve (AUC). The AUC is particularly prioritized as it provides a robust measure of discriminative ability at various threshold settings, independent of class distribution.

\textbf{Survival Analysis Metrics.} For prognosis prediction, we employ the Concordance Index (C-index) as the primary evaluation metric. The C-index measures the fraction of all pairs of subjects whose predicted survival times are correctly ordered relative to the true survival times. A C-index of 0.5 indicates random prediction, while 1.0 indicates perfect ranking capability.

\textbf{Statistical Analysis.} To determine the statistical significance of the performance improvements, we adopt the paired t-test. In our comparative analysis, if the calculated p-value between two results is larger than 0.05, the difference is considered statistically insignificant, indicating that the two results are comparable.

\subsection{Comparison with State-of-the-Art Methods}
To comprehensively assess the effectiveness and generalization capability of the proposed framework, we conducted systematic experiments covering a broad spectrum of tasks, including cancer subtyping, tumor diagnosis (cancer vs. normal), and survival risk stratification. 

We benchmarked our method against diverse representative baselines covering various MIL paradigms: (1) Conventional pooling-based methods: Mean-Pooling and Max-Pooling MIL \citep{campanella2019clinical}; (2) Attention-based MIL: The vanilla ABMIL \citep{ilse2018attention}, along with its advanced variants DSMIL \citep{li2021dual} (utilizing non-local attention) and CLAM \citep{lu2021data} (including both single-branch CLAM-SB and multi-branch CLAM-MB); (3) Transformer-based MIL: TransMIL \citep{shao2021transmil}, designed to capture long-range dependencies; (4) Feature optimization methods: The pseudo-bag augmented DTFD-MIL \citep{zhang2022dtfd}; (5) Recent SOTA frameworks: The vision-language aligned ViLa-MIL \citep{shi2024vila} and the Shapley value-augmented PMIL \citep{yan2025shapley}; (6) Survival-specific frameworks: For survival analysis tasks, we include representative frameworks to ensure a rigorous comparison: ProSurv \citep{liu2025prototype}, a specialized prototype-guided framework; ProtoSurv \citep{NEURIPS2024_760341ad}, a heterogeneous graph learning model that explicitly incorporates pathological domain knowledge to capture intratumoral heterogeneity and spatial interactions; and DM-GNN \citep{WANG2024103252}, a dual-stream architecture that models whole slide images via graphs based on morphological affinity.

\subsection{Classification Performance}
Table \ref{tab2} presents the quantitative comparison across four datasets. The proposed HPDP framework consistently achieves state-of-the-art performance in terms of ACC, AUC, and F1-score, outperforming both conventional single-modal baselines and recent multimodal approaches. While attention-based methods and Transformer-based TransMIL demonstrate reasonable performance on standard benchmarks, they exhibit significant instability on the heterogeneous, in-house LCEM dataset (e.g., TransMIL drops to 54.85\% AUC). Furthermore, although advanced frameworks incorporating semantic guidance or feature attribution mechanisms achieve competitive second-best results, HPDP still demonstrates a clear advantage by effectively integrating structured domain knowledge.

The robustness of our method is particularly evident in challenging scenarios. On the internal LCEM cohort for EGFR mutation prediction, HPDP attains an AUC of 83.64\%, surpassing the runner-up PMIL (82.29\%) and ViLa-MIL (81.41\%). Similarly, for fine-grained histological subtyping on BRACS, HPDP achieves a leading F1-score of 65.61\%, significantly outperforming CLAM (62.58\%). This superior performance is primarily attributed to the integration of domain priors and the MAPS, which effectively disentangles task-specific morphological patterns from background noise. Complementing this structural backbone, the cross-modal alignment facilitated by the HCMA module further reinforces semantic consistency, jointly yielding a representation that is more discriminative than purely data-driven approaches.

\begin{table*}
\centering
\caption{Comparison of WSI classification performance (mean \% $\pm$ std). \textbf{Bold} indicates the best result, an \underline{underline} indicates the second best, and * denotes comparable performance to the top result (paired t-test, p $>$ 0.05).}
\resizebox{\textwidth}{!}{
\begin{tabular}{ccccccccccccc}
\toprule
\textbf{Method} & \multicolumn{3}{c}{\textbf{\text{LCEM}}} & \multicolumn{3}{c}{\textbf{Camelyon16}} & \multicolumn{3}{c}{\textbf{TCGA-Lung}} & \multicolumn{3}{c}{\textbf{BRACS}}\\
\cmidrule(r){2-4} \cmidrule(r){5-7} \cmidrule(r){8-10} \cmidrule(r){11-13}
& ACC & AUC & F1 & ACC & AUC & F1 & ACC & AUC & F1 & ACC & AUC & F1\\
\midrule
MeanPooling & 62.21{\scriptsize$\pm$6.82} & 62.11{\scriptsize$\pm$7.00} &42.60{\scriptsize$\pm$11.69} & 74.23{\scriptsize$\pm$8.12} & 75.56{\scriptsize$\pm$9.33} & 72.14{\scriptsize$\pm$8.91} & 76.32{\scriptsize$\pm$3.71} & 83.56{\scriptsize$\pm$3.34} & 76.25{\scriptsize$\pm$3.68} & 65.26{\scriptsize$\pm$1.31} & 79.78{\scriptsize$\pm$2.23} & 46.30{\scriptsize$\pm$1.74}\\
MaxPooling & 64.36{\scriptsize$\pm$7.42} & 66.68{\scriptsize$\pm$9.78} & 46.89{\scriptsize$\pm$14.73}
               &73.78{\scriptsize$\pm$9.66} & 75.34{\scriptsize$\pm$9.88} & 72.07{\scriptsize$\pm$7.67} & 80.57{\scriptsize$\pm$2.46} & 89.61{\scriptsize$\pm$1.78} & 80.52{\scriptsize$\pm$2.51} & 66.37{\scriptsize$\pm$3.36} & 82.01{\scriptsize$\pm$2.21} & 51.33{\scriptsize$\pm$6.46}\\
CLAM & 68.74{\scriptsize$\pm$7.23} & 70.27{\scriptsize$\pm$11.68} & 60.09{\scriptsize$\pm$15.06}& 80.78{\scriptsize$\pm$4.73} & 83.04{\scriptsize$\pm$4.48} & 78.29{\scriptsize$\pm$5.63} & 85.94{\scriptsize$\pm$1.08} & 93.08{\scriptsize$\pm$1.22} & 85.87{\scriptsize$\pm$1.05} & \underline{75.50{\scriptsize$\pm$1.27}}* & 88.02{\scriptsize$\pm$0.95}* & 62.58{\scriptsize$\pm$5.97}\\
ABMIL & 66.75{\scriptsize$\pm$9.75} & 69.00{\scriptsize$\pm$8.22} & 54.80{\scriptsize$\pm$17.19}& 76.74{\scriptsize$\pm$9.07} & 78.36{\scriptsize$\pm$7.96} & 74.37{\scriptsize$\pm$18.00} & 85.47{\scriptsize$\pm$1.47} & \underline{93.80{\scriptsize$\pm$1.54}} & 85.35{\scriptsize$\pm$1.51} & 74.22{\scriptsize$\pm$2.00}* & 88.06{\scriptsize$\pm$2.17}* & 59.40{\scriptsize$\pm$7.14}\\
DSMIL & 68.28{\scriptsize$\pm$3.76} & 69.75{\scriptsize$\pm$4.20} & 63.37{\scriptsize$\pm$4.36} 
               & 75.35{\scriptsize$\pm$6.31} & 78.26{\scriptsize$\pm$6.75} & 73.41{\scriptsize$\pm$8.95} & 83.02{\scriptsize$\pm$2.89} & 90.23{\scriptsize$\pm$2.31} & 83.01{\scriptsize$\pm$2.89} & 70.75{\scriptsize$\pm$1.76} & 86.53{\scriptsize$\pm$1.57}* & 55.88{\scriptsize$\pm$3.83}\\
TransMIL  &  60.40{\scriptsize$\pm$4.82} & 54.85{\scriptsize$\pm$5.15} & 37.61{\scriptsize$\pm$1.93} & 78.33{\scriptsize$\pm$7.23} & 79.25{\scriptsize$\pm$8.17} & 77.92{\scriptsize$\pm$9.87} & 80.00{\scriptsize$\pm$6.98} & 88.14{\scriptsize$\pm$8.51} & 79.91{\scriptsize$\pm$7.00} & 69.30{\scriptsize$\pm$5.42} & 82.87{\scriptsize$\pm$3.83} & 52.74{\scriptsize$\pm$5.21}\\
DTFD-MIL  &  61.25{\scriptsize$\pm$8.00} & 62.68{\scriptsize$\pm$7.56} & 62.93{\scriptsize$\pm$12.36} & 83.57{\scriptsize$\pm$2.27}* & 86.36{\scriptsize$\pm$1.87} & 75.18{\scriptsize$\pm$3.37} & \underline{88.11{\scriptsize$\pm$1.08}}* & 93.54 {\scriptsize$\pm$1.32} & \underline{87.43 {\scriptsize$\pm$1.06}}* & 71.11{\scriptsize$\pm$4.06} & 83.71{\scriptsize$\pm$1.46} & 64.36{\scriptsize$\pm$4.92}*  \\
ViLa-MIL  & \underline{75.58{\scriptsize$\pm$3.14}}* & 81.41{\scriptsize$\pm$4.04}* & \underline{72.57{\scriptsize$\pm$3.65}}* & 82.79{\scriptsize$\pm$1.93}* & 86.60{\scriptsize$\pm$3.01} & 81.27{\scriptsize$\pm$1.90} & 87.26{\scriptsize$\pm$1.97}* & 92.80{\scriptsize$\pm$1.18} & 87.26{\scriptsize$\pm$1.97}* & 74.31{\scriptsize$\pm$2.77}* & \underline{88.13{\scriptsize$\pm$2.15}}* & \underline{64.38{\scriptsize$\pm$6.90}}*\\
PMIL & 73.16{\scriptsize$\pm$1.79}* & \underline{82.29{\scriptsize$\pm$1.84}}* & 72.27{\scriptsize$\pm$1.87}* & \underline{83.88{\scriptsize$\pm$0.90}}* & \underline{87.27{\scriptsize$\pm$2.36}}* & \underline{81.84{\scriptsize$\pm$1.60}}* & 85.00{\scriptsize$\pm$1.85} & 91.50{\scriptsize$\pm$1.24} & 84.96{\scriptsize$\pm$1.89} & 72.03{\scriptsize$\pm$1.81}* & 87.24{\scriptsize$\pm$1.72}* & 58.74{\scriptsize$\pm$7.13}\\
\midrule
\textbf{Ours} & \textbf{76.12{\scriptsize$\pm$1.21}} & \textbf{83.64{\scriptsize$\pm$1.08}} & \textbf{73.44{\scriptsize$\pm$1.56}} & \textbf{85.17{\scriptsize$\pm$0.92}} & \textbf{89.05{\scriptsize$\pm$1.87}} & \textbf{82.90{\scriptsize$\pm$1.35}} & \textbf{89.34{\scriptsize$\pm$2.37}} & \textbf{95.61{\scriptsize$\pm$2.47}} & \textbf{89.31{\scriptsize$\pm$2.39}} & \textbf{76.35{\scriptsize$\pm$1.69}} & \textbf{89.43{\scriptsize$\pm$1.59}} & \textbf{65.61{\scriptsize$\pm$6.05}}\\
\bottomrule
\end{tabular}}
\label{tab2}
\end{table*}

\subsection{Survival Analysis Performance}

To validate the prognostic value of the proposed framework, we evaluated the performance of HPDP on four independent cancer cohorts. As summarized in Table \ref{tab3}, the model yields an overall average C-index of 0.6577 across all datasets. Specifically, on the in-house LCEM cohort, HPDP achieves a C-index of 0.6985 with a standard deviation of 0.0317. In comparison, the graph-based method ProSurv recorded a C-index of 0.6797, and the prototype-based ProtoSurv achieved 0.6543. These quantitative results indicate the performance of the proposed method on clinical data relative to existing baselines.

Extending the evaluation to public benchmarks, HPDP records C-indices of 0.6941 on TCGA-COAD, 0.5980 on TCGA-LUAD, and 0.6401 on TCGA-PAAD. These outcomes are statistically comparable to top-performing methods. Notably, while DM-GNN utilizes explicit graph modeling and ProSurv employs cross-modal genomic enhancement, our results indicate that prognostic stratification performance comparable to graph-based or genome-enhanced approaches can be achieved through the integration of structured visual prototypes, geometric spatial encoding, and text-guided feature alignment.

\begin{table*}[htbp]
    \centering
    \caption{Comparison of survival analysis performance (C-index, mean $\pm$ std). \textbf{Bold} indicates the best result, and \underline{underline} indicates the second best, and * denotes comparable performance to the top result (paired t-test, p $>$ 0.05).}
    \label{tab3}

    \begin{tabular}{lccccc}
        \toprule
        \textbf{Methods} & \textbf{LCEM} & \textbf{TCGA-LUAD} & \textbf{TCGA-COAD} & \textbf{TCGA-PAAD} & \textbf{Overall}\\
        \midrule
        MaxPooling & 0.5019$ \pm $0.0751 & 0.5610$ \pm $0.0262 & 0.5683$ \pm $0.0160 & 0.5433$ \pm $0.0235 & 0.5436\\
        MeanPooling & 0.5139$ \pm $0.0580 & 0.5594$ \pm $0.0403 & 0.5557$ \pm $0.0385 & 0.5518$ \pm $0.0483 & 0.5452\\
        ABMIL & 0.6195$ \pm $0.1478 & 0.5449$ \pm $0.0168 & 0.6218$ \pm $0.0262 & 0.6112$ \pm $0.0384 & 0.5994\\
        CLAM-SB & 0.5668$ \pm $0.1077 & 0.5755$ \pm $0.0507 & 0.6496$ \pm $0.0196 & 0.6019$ \pm $0.0464 & 0.5985\\
        DSMIL & 0.5699$ \pm $0.0810 & 0.5247$ \pm $0.0485 & 0.6396$ \pm $0.0397 & 0.6223$ \pm $0.0310 & 0.5891\\
        TransMIL & 0.6211$ \pm $0.0614 & 0.5372$ \pm $0.0236 & 0.6431$ \pm $0.0470 & 0.6190$ \pm $0.0569 & 0.6051\\
        DM-GNN & 0.6433$ \pm $0.0528 & 0.5784$ \pm $0.0295 & 0.6780$ \pm $0.0569* & 0.6315$ \pm $0.0391 & 0.6328\\
        ProtoSurv & 0.6543$ \pm $0.0423 & 0.5657$ \pm $0.0310 & 0.6725$ \pm $0.0624* & \textbf{0.6453$ \pm $0.0512} & 0.6345\\
        ProSurv & \underline{0.6797$ \pm $0.0446*} & \textbf{0.5982$ \pm $0.0268} & \underline{0.6833$ \pm $0.0510}* & \underline{0.6413$ \pm $0.0435}* & \underline{0.6506}\\
        \midrule
        \textbf{Ours} & \textbf{0.6985 $\pm$ 0.0317} & \underline{0.5980$ \pm $0.0286}* & \textbf{0.6941$ \pm $0.0461} & 0.6401$ \pm $0.0410* & \textbf{0.6577}\\
        \bottomrule
    \end{tabular}
\end{table*}

\subsection{Zero-Shot Generalization Performance}

To evaluate robustness under domain shifts, we conducted zero-shot experiments by training on LCEM and testing directly on TCGA-LUAD. Given the extreme class imbalance (Wild-type $\gg$ Mutant) in the target domain, we prioritize AUC and F1-score as primary metrics. In classification, HPDP achieves an AUC of 75.50\% and F1-score of 65.47\%, significantly surpassing the runner-up PMIL (72.30\%). In contrast, TransMIL exhibits severe mode collapse, tending to classify all samples as the majority Wild-type class. This trivial prediction strategy results in a near-random AUC (52.47\%) and a critically low F1-score (12.29\%), highlighting its failure to extract domain-invariant features. Conversely, the stability of HPDP suggests that MAPS effectively captures robust morphological structures rather than overfitting to class priors.

For survival prediction, our framework yields a C-index of 0.5954, outperforming both the graph-based DM-GNN (0.5594) and the specialized survival model ProSurv (0.5652). These results confirm that integrating explicit geometric encoding (SPE) and semantic alignment (HCMA) enables the model to learn universal prognostic biomarkers that are transferable across different patient populations.

\begin{table}[t]
    \centering
    \caption{Zero-shot generalization performance. Models are trained on the internal LCEM cohort (Source) and directly tested on the external TCGA-LUAD cohort (Target) without any fine-tuning. We report AUC for tumor subtype classification (Wild-type vs. Mutant) and C-index for survival risk prediction. \textbf{Notably, given the extreme class imbalance of EGFR genotypes in the TCGA-LUAD cohort, we prioritize AUC and F1-score as the primary metrics for robust evaluation.} Bold indicates the best result, \underline{underline} indicates the second best, and * denotes comparable performance to the top result (paired t-test, $p > 0.05$).}
    \label{tab:zero_shot}
    
    \resizebox{\columnwidth}{!}{
    \begin{tabular}{l|ccc|c}
        \toprule

        Method &\textbf{ACC (\%)}& \textbf{AUC (\%)} & \textbf{F1 (\%)} & \textbf{C-index} \\
        \midrule
        MaxPooling & 32.58$ \pm $25.59 & 57.31$ \pm $1.94 & 27.58$ \pm $20.96 & 0.5217$ \pm $0.0327 \\
        MeanPooling & 24.13$ \pm $22.61 & 58.94$ \pm $1.80 & 20.41$ \pm $18.16 & 0.5290$ \pm $0.0221\\
        ABMIL       & 41.17$ \pm $24.86 & 59.93$ \pm $2.06 & 34.77$ \pm $20.54 & 0.5388$ \pm $0.0357 \\
        CLAM-SB     & 57.08$ \pm $24.75 & 60.10$ \pm $3.21 & 44.09$ \pm $12.52 & 0.5161$ \pm $0.0470 \\
        DSMIL & 62.20$ \pm $8.57 & 60.34$ \pm $0.78 & 52.88$ \pm $1.75 & 0.5333$ \pm $0.0195\\
        TransMIL & 14.02$ \pm $0.00 & 52.47$ \pm $5.29 & 12.29$ \pm $0.00 & 0.5012$ \pm $0.0334\\
        ProtoSurv & - & - & - & 0.5453$ \pm $0.0612\\
        DM-GNN & - & - & - & 0.5594$ \pm $0.0508\\
        ViLa-MIL    & 55.34$\pm$27.17 & 52.59$\pm$5.46 &  39.66$\pm$13.00              & -     \\
        ProSurv     & -                & -                & - & \underline{0.5652$\pm$0.0527} \\
        PMIL & \underline{65.27$\pm$8.49}* & \underline{72.30$\pm$2.25}* & \underline{60.03$\pm$4.21}* & - \\
        \midrule

        \textbf{HPDP (Ours)} & \textbf{70.85$\pm$6.31} & \textbf{75.50$\pm$3.98} & \textbf{65.47$\pm$3.44} & \textbf{0.5954$\pm$0.0445} \\
        \bottomrule
    \end{tabular}}
\end{table}

\subsection{Qualitative Results and Visualization}

To comprehensively evaluate the HPDP framework, we conducted a multi-dimensional analysis spanning spatial interpretability, feature discriminability, and prognostic value.

\begin{figure*}[t]
  \centering
  \includegraphics[width=\linewidth]{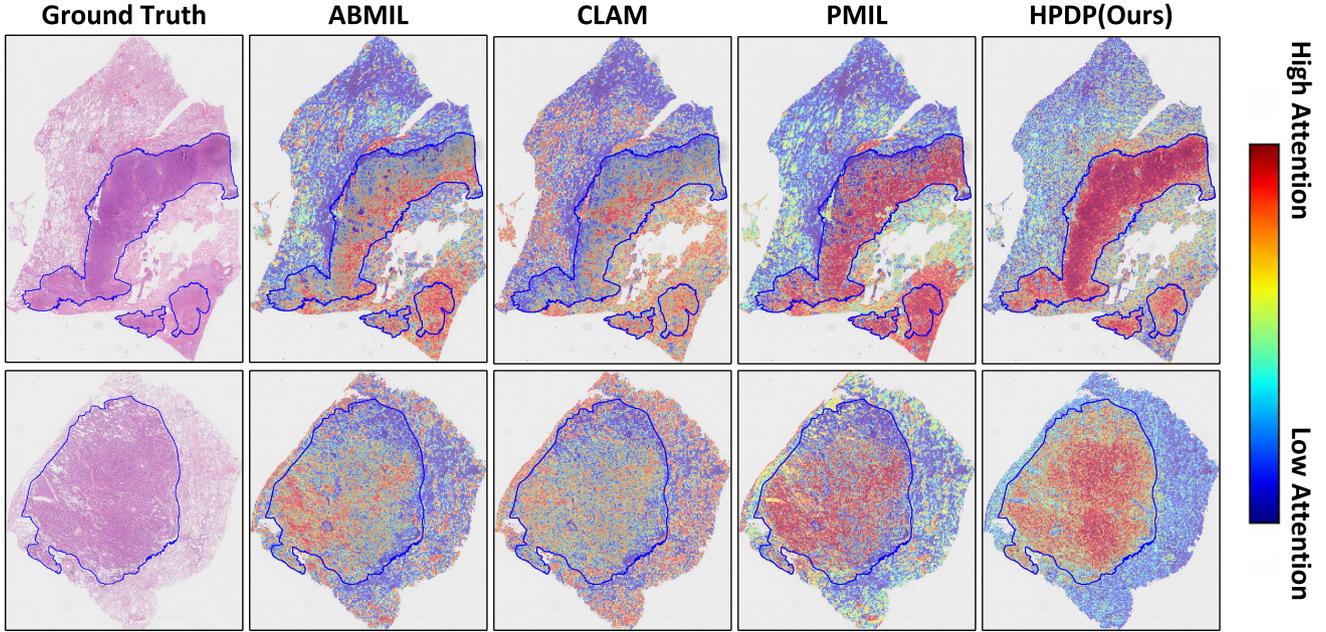}
  \caption{Visualization of spatial attention heatmaps compared with state-of-the-art methods. The first column displays the original WSI with pathologist-annotated tumor regions (blue contours). Subsequent columns illustrate attention maps from ABMIL, CLAM, PMIL, and HPDP. }
  \label{fig:heatmap}
\end{figure*}

\textbf{Spatial Interpretability via Attention Heatmaps.} Visualizing spatial attention distributions (Fig. \ref{fig:heatmap}) provides direct insight into the model's decision-making focus. Standard attention-based benchmarks often suffer from "attention scattering," where the model erroneously highlights irrelevant stroma or background regions due to overfitting on spurious correlations. In contrast, HPDP demonstrates exceptional spatial coherence. Its high-attention regions align precisely with tumor morphology and remain strictly confined within ground truth boundaries, confirming that our semantic guidance successfully steers the model toward biologically meaningful entities.

\begin{figure*}[t]
  \centering
  \includegraphics[width=\linewidth]{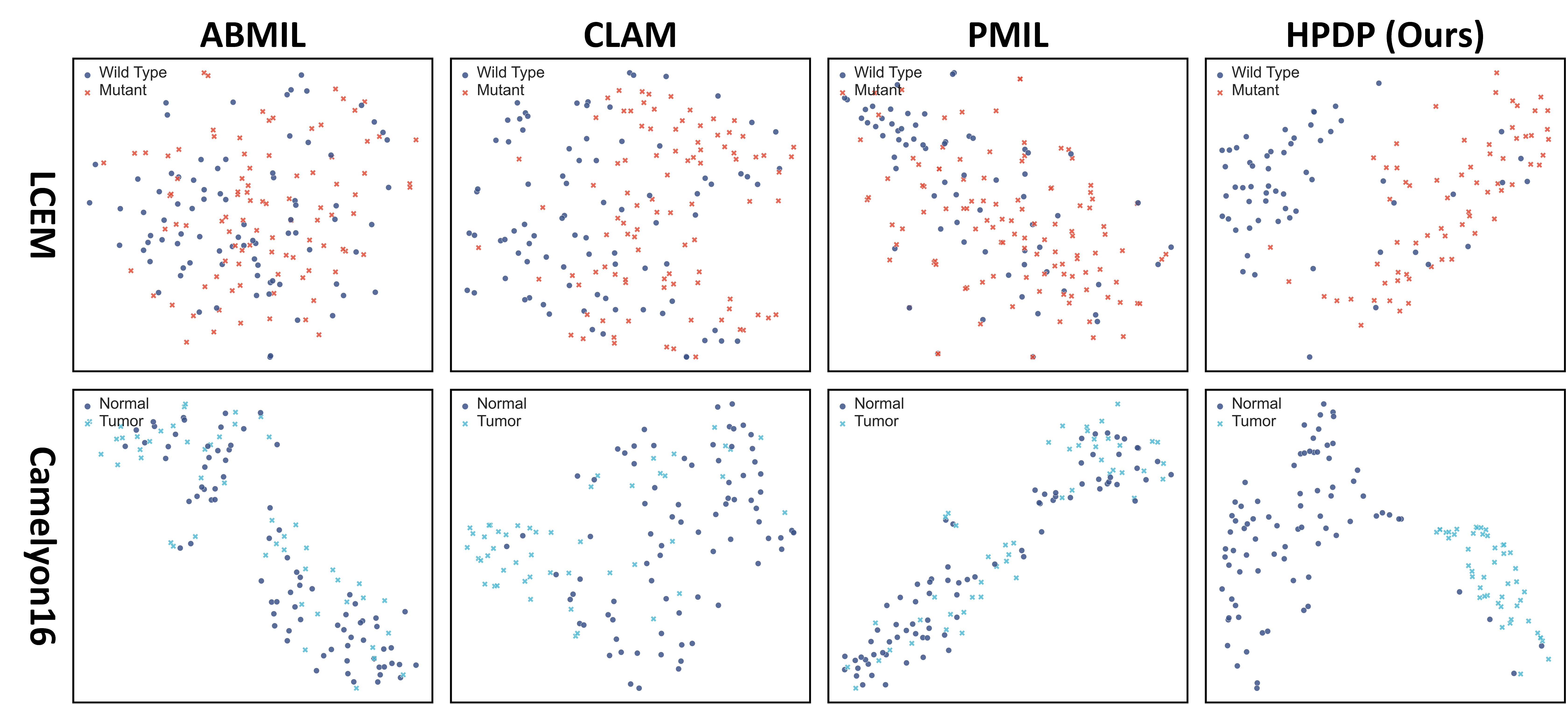}
  \caption{t-SNE visualization of slide-level feature representations on LCEM (Top) and Camelyon16 (Bottom). The top row illustrates the subtype classification task on LCEM, while the bottom row shows metastasis detection on Camelyon16. Columns from left to right display feature spaces learned by ABMIL, CLAM, PMIL, and HPDP.}
  \label{fig:tsne}
\end{figure*}

\textbf{Feature Space Discriminability.} Beyond local interpretability, the quality of the latent feature space is assessed using t-SNE \citep{van2008visualizing} visualizations (Fig. \ref{fig:tsne}). While existing MIL benchmarks exhibit significant feature entanglement, particularly in the challenging LCEM subtype classification task, HPDP generates highly compact and well-separated clusters. This superior separability indicates that the integration of morphological priors effectively reduces intra-class variance, yielding robust representations that are sensitive to subtle decision boundaries.

\begin{figure*}[t]
  \centering
  \includegraphics[width=\linewidth]{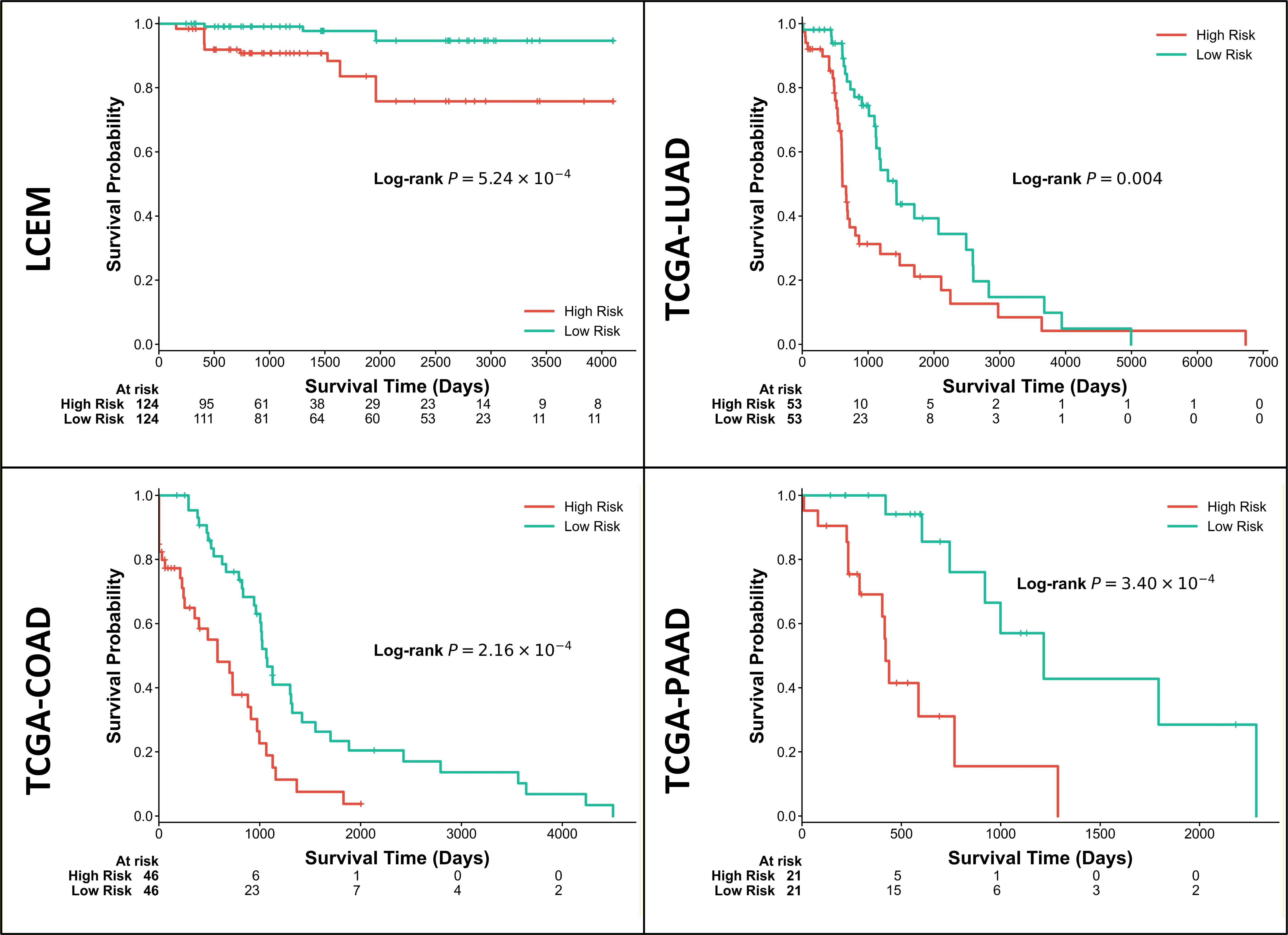}
  \caption{Kaplan-Meier survival curves of HPDP across four independent cohorts. Patients are stratified into High Risk (red) and Low Risk (green) groups based on the median risk score. The Log-rank test indicates statistically significant stratification in all datasets.}
  \label{fig:km_curves}
\end{figure*}

\textbf{Prognostic Stratification.} The clinical validity of these learned features is further corroborated by Kaplan-Meier survival analysis across four independent cohorts (Fig. \ref{fig:km_curves}). Our framework achieves statistically significant risk stratification across all datasets ($p < 0.005$). Notably, in the aggressive phenotypes of TCGA-COAD ($p=2.16 \times 10^{-4}$) and TCGA-PAAD ($p=3.40 \times 10^{-4}$), the model exhibits a decisive ability to distinguish patient risks, demonstrating its robustness in capturing survival-associated biomarkers.

\subsection{Ablation Study}

To explicitly quantify the contribution of different input modalities and the effectiveness of key components, we conducted a step-wise ablation study as summarized in Table \ref{tab:llm_ablation}. The uni-modal vision baseline yields limited performance. Upon integrating clinical text descriptions, a performance increase of +10.48\% AUC was observed, suggesting the potential benefit of semantic guidance in complementing visual features. Regarding specific modules, the inclusion of the MAPS coincided with a decrease in performance variance (std from $\pm$ 6.05 to $\pm$ 4.13) and an improvement in classification auc. Notably, the configuration incorporating only HCMA achieved the highest C-index (0.6003) on TCGA-LUAD. We attribute the full model's marginal drop to the use of unified hyperparameters; while this fixed setting is optimal for LCEM and prevents overfitting, it causes minor fluctuations on the TCGA-LUAD cohort. Ultimately, the comprehensive integration of morphological priors via MAPS, semantic alignment via HCMA, and geometric constraints via SPE yielded the highest overall performance (0.6985 C-index on LCEM), indicating that these modules function effectively in concert to support the final prediction.

In addition to component-wise analysis, we investigated the impact of the number of Prior Experts ($\mathcal{K}_{prior}$) of MAPS. As presented in Table \ref{tab:ablation_k}, although $\mathcal{K}_{prior}=3$ yields a marginally higher C-index on TCGA-LUAD, the model achieves its peak performance on both LCEM and Camelyon16 when $\mathcal{K}_{prior}=4$. This empirical optimum likely reflects the four canonical tissue phenotypes in the tumor microenvironment, where $\mathcal{K}=4$ allows the MAPS module to effectively capture these distinct entities without the under-segmentation risks of smaller $\mathcal{K}$ or the redundancy introduced by larger values. Consequently, $\mathcal{K}_{prior}=4$ was selected as the default configuration to ensure the best overall balance between interpretability and performance.

\begin{table*}[htbp]
    \centering
    \caption{Step-wise ablation study investigating the contribution of different input modalities and the effectiveness of key components in the proposed HPDP framework. \textbf{SPE}: Sinusoidal Positional Encoder; \textbf{MAPS}: Morphologically Anchored Prototype System; \textbf{HCMA}: Hierarchical Cross-Modal Alignment. \textbf{Bold} indicates the best performance.}
    \label{tab:llm_ablation}
  
    \begin{tabular}{cccccc|cc|cc}
        \toprule
        \multicolumn{6}{c|}{Input Modalities \& Key Components} & \multicolumn{2}{c|}{Classification (AUC \%)} & \multicolumn{2}{c}{Survival (C-index)} \\
        \cmidrule(r){1-6} \cmidrule(lr){7-8} \cmidrule(l){9-10}
        Vision & Text & Coord & MAPS &  HCMA & SPE &  LCEM & Camelyon16 & LCEM & TCGA-LUAD \\
        \midrule
        \checkmark &  & & & && 69.00$\pm$8.22 & 78.36$\pm$7.96 & 0.6195$ \pm $0.1478 & 0.5449$ \pm $0.0168 \\
        \checkmark &\checkmark & & &  & & 79.48$\pm$5.43  & 85.21$\pm$5.63 & 0.6450$ \pm $0.0812 &  0.5661$ \pm $0.0410\\

        \checkmark &\checkmark &\checkmark & &  &  & 80.25$\pm$6.05  & 85.44$\pm$6.12 & 0.6499$ \pm $0.0732 & 0.5548$ \pm $0.0515 \\

        \checkmark &\checkmark &\checkmark &\checkmark &  & & 82.80$\pm$4.13  & 87.92$\pm$3.27 & 0.6924$ \pm $0.0379 & 0.5875$ \pm $0.0342 \\

        \checkmark &\checkmark &\checkmark & &\checkmark & & 82.43$\pm$1.44 & 88.90$\pm$1.82 & 0.6980$ \pm $0.0288 &  \textbf{0.6003$ \pm $0.0233}\\

        \checkmark & \checkmark &\checkmark & &  &\checkmark & 80.95$\pm$4.38 & 86.74$\pm$4.12  & 0.6711$ \pm $0.0443 & 0.5742$ \pm $0.0380\\

        \checkmark & \checkmark &\checkmark & \checkmark & \checkmark & \checkmark & \textbf{83.64$\pm$1.08} & \textbf{89.05$\pm$1.87} & \textbf{0.6985$\pm$0.0317} & 0.5980$\pm$0.0286 \\
        \bottomrule
    \end{tabular}
\end{table*}

\begin{table}[htbp]
    \centering
    \caption{Ablation study on the number of Prior Experts ($\mathcal{K}_{prior}$) in the MAPS. We investigate the impact of varying the granularity of domain priors on both classification and survival tasks. \textbf{Bold} indicates the best performance.}
    \label{tab:ablation_k}

    \resizebox{\columnwidth}{!}{
        \begin{tabular}{c|cc|cc}
            \toprule
            \multirow{2}{*}{$\mathcal{K}_{prior}$} & \multicolumn{2}{c|}{Classification (AUC \%)} & \multicolumn{2}{c}{Survival (C-index)} \\
            \cmidrule(lr){2-3} \cmidrule(l){4-5}
             & LCEM & Camelyon16 & LCEM & TCGA-LUAD \\
            \midrule
            3 & 82.07$\pm$2.36 & 88.94$\pm$2.56 & 0.6731$\pm$0.0428 & \textbf{0.5992$\pm$0.0273} \\
            4 & \textbf{83.64$\pm$1.08} & \textbf{89.05$\pm$1.87} & \textbf{0.6985$\pm$0.0317} & 0.5980$\pm$0.0286\\
            5 & 83.15$\pm$0.79 & 86.43$\pm$1.96 & 0.6512$\pm$0.0431 & 0.5684$\pm$0.0364 \\
            6 & 81.60$\pm$2.45 & 85.47$\pm$3.01 & 0.6795$\pm$0.0287 & 0.5790$\pm$0.0295 \\
            \bottomrule
        \end{tabular}
    }
\end{table}

\begin{figure*}
    \centering
    \includegraphics[width=1\linewidth]{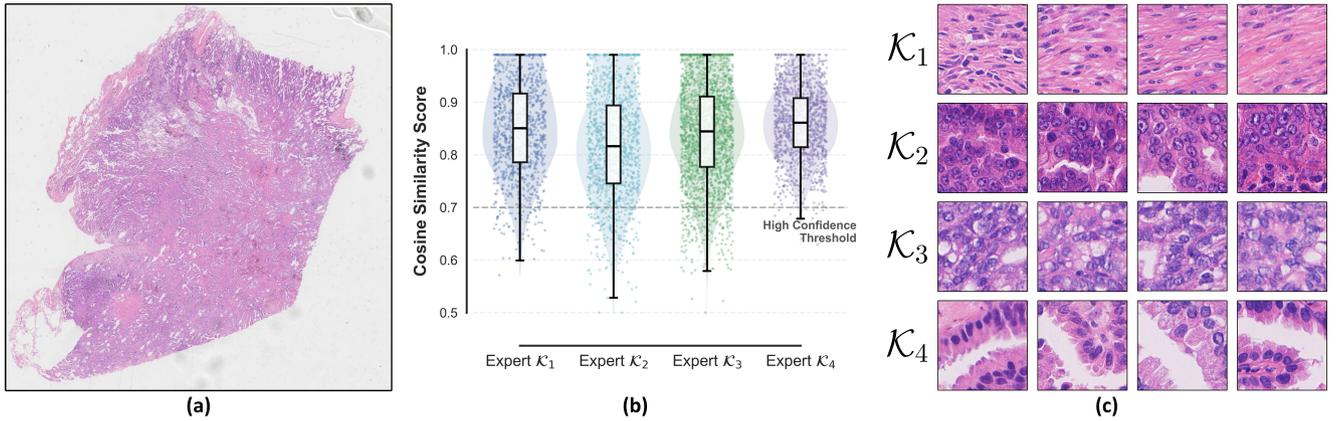}
    \caption{Qualitative and quantitative visualization of the MAPS.
(a) A representative WSI from the LCEM cohort. (b) Quantitative analysis of cosine similarity scores. The violin plots display the distribution of similarity scores for the learned experts ($\mathcal{K}_1$–$\mathcal{K}_4$), showing the density of samples relative to the 0.7 threshold. (c) Top-4 retrieved patches for each expert.}
    \label{fig6}
\end{figure*}

\section{Discussion}

In this study, we presented HPDP, a multimodal framework designed to address the twin challenges of heterogeneity and data scarcity in pathological survival analysis. By integrating MAPS, SPE, and HCMA, our method effectively bridges the gap between low-level visual signals and high-level clinical semantics. Extensive experiments on seven independent cohorts demonstrate that HPDP yields robust prognostic stratification and subtype classification performance, statistically comparable to or surpassing state-of-the-art graph-based and genome-enhanced approaches.

\subsection{Interpretability of Morphological Prototypes}
A key distinguishing feature of our approach is the explicit structuring of the feature space via MAPS. Unlike conventional MIL methods that operate in a latent space often detached from biological reality, MAPS enforces a correspondence between learnable prototypes and canonical tissue phenotypes. As observed in our ablation studies, setting the number of prototypes $\mathcal{K}_{prior}=4$ yields optimal performance, which aligns with the fundamental histological components of the tumor microenvironment. This biological consistency is further corroborated by the visualization in Fig. \ref{fig6}, where the learned experts spontaneously disentangle distinct morphological patterns with high confidence scores. This evidence suggests that the model does not merely learn statistical correlations but effectively identifies and utilizes biologically meaningful structures to formulate its predictions \citep{fan2025ipmmg,krishna2023interpretable}.

\subsection{Robustness Under Domain Shifts}
Generalization across institutions is a bottleneck for clinical deployment due to variations in staining protocols and scanner specifications \citep{vorontsov2024foundation,campanella2025real}. Our zero-shot experiments on TCGA-LUAD highlight a critical advantage of the proposed framework. While standard attention-based models suffered from mode collapse by gravitating towards the majority class due to stain variations, HPDP maintained high discriminability (AUC 75.50\%). We attribute this robustness to the synergistic effect of SPE and HCMA. The geometric constraints provided by SPE ensure that spatial tissue architecture is preserved regardless of color shifts, while the text-guided alignment in HCMA anchors visual features to invariant semantic concepts rather than site-specific pixel statistics.

\subsection{Clinical Implications}
From a clinical perspective, the proposed vision-language alignment offers a step towards "white-box" diagnostic support \citep{chen2024camil}. By modulating visual attention with clinical text descriptions, HPDP simulates the diagnostic reasoning process of a pathologist, who interprets visual patterns in the context of clinical knowledge. This capability is particularly valuable in survival analysis, where the model can identify risk-associated regions that are consistent with established medical literature, thereby enhancing trust and facilitating adoption in clinical workflows \citep{khanal2025hallucination}.

\subsection{Limitations and Future Work}
Despite promising results, limitations remain. First, reliance on LLM-generated descriptions introduces a dependency on generative fidelity, where potential hallucinations could propagate noise into the HCMA module. Second, a domain gap exists between structured synthesized text and the noisy, unstructured EHRs in routine practice. Future work will move beyond statistical alignment to explore causal associations between textual semantics and visual patterns. We will investigate causal intervention techniques to distinguish intrinsic pathological evidence from spurious correlations, thereby enhancing robustness on real-world clinical data \citep{campanella2025real,Arostegui2026Challenges}.

\section{Conclusion}

In this work, we introduced HPDP to address the semantic disconnect and lack of spatial awareness in multimodal histopathology analysis. Through the synergistic integration of MAPS, SPE, and HCMA, our framework achieves state-of-the-art accuracy across seven cohorts and demonstrates significant zero-shot robustness under domain shifts. By explicitly anchoring visual features to biological phenotypes and preserving tissue geometry, HPDP moves beyond the opaque 'black-box' paradigm, offering a transparent and clinically grounded decision process. These results suggest that our approach can serve as a reliable and interpretable tool for precision oncology.

\section*{CRediT authorship contribution statement}

\textbf{Xuemei Qiu}: Methodology, Data curation, Validation, Software, Formal analysis, Writing - original draft, Visualization.   
\textbf{Dawei Fan}: Conceptualization, Data curation, Software, Writing - review \& editing. 
\textbf{Yebin Huang}: Validation, Writing - review \& editing. 
\textbf{Yanping Chen}:  Supervision, Resources, Validation.
\textbf{Lifang Wei}: Conceptualization, Formal analysis, Resources, Funding acquisition, Project administration, Writing - review \& editing.

\section*{Declaration of competing interest}
The authors declare that they have no known competing financial interests or personal relationships that could have appeared to influence the work reported in this paper.

\section*{Data Availability}

The datasets analyzed in this study are available from the following public repositories: The TCGA datasets can be accessed via the GDC Data Portal (\url{https://portal.gdc.cancer.gov/}). The Camelyon16 dataset is available at \url{https://camelyon16.grand-challenge.org/}. The BRACS dataset is publicly available at \url{https://www.bracs.icar.cnr.it/}.

The in-house dataset used for development and validation was retrospectively collected from a medical center in China. Due to patient privacy regulations and institutional policies, these proprietary clinical data are not publicly available. However, de-identified data may be made available by the corresponding author upon reasonable request, subject to the provision of necessary local and national ethical approvals by the requesting party.

\section*{Acknowledgments}
This work was supported in part by National Natural Science Foundation of China (Grant No.62571127, 62171130, 62301160), the Natural Science Foundation of Fujian Province (Grant No.2025Y0066, 2025Y0069), and Fujian Health Science and Technology Plan Project (Grant No.2024CXA035).

\bibliographystyle{cas-model2-names}

\bibliography{cas-refs}

\end{document}